\documentclass[sigconf]{acmart}

\AtBeginDocument{%
  }

\usepackage{ccicons}

\copyrightyear{2025}
\acmYear{2025}
\setcopyright{rightsretained}

\acmConference[WWW '25]{Proceedings of the ACM Web Conference 2025}{April 28-May 2, 2025}{Sydney, NSW, Australia} \acmBooktitle{Proceedings of the ACM Web Conference 2025 (WWW '25), April 28-May 2, 2025, Sydney, NSW, Australia} \acmDOI{10.1145/3696410.3714526} \acmISBN{979-8-4007-1274-6/25/04}

\makeatletter
\gdef\@copyrightpermission{
 \begin{minipage}{0.2\columnwidth}
  \href{https://creativecommons.org/licenses/by-nc/4.0/}{\ccbync}
 \end{minipage}\hfill
 \begin{minipage}{0.8\columnwidth}
  \href{https://creativecommons.org/licenses/by-nc/4.0/}{This work is licensed under a Creative Commons Attribution-NonCommercial International 4.0 License.}
 \end{minipage}
 \vspace{5pt}
}
\makeatother

\usepackage{graphicx}
\usepackage{booktabs}
\usepackage{supertabular}
\usepackage{todonotes}
\usepackage{pgfplots}
\pgfplotsset{compat=1.18}
\usepackage{lipsum}
\usepackage{balance}
\usepackage{float}
\usepackage{placeins}    
\usepackage{tabularx}

\begin{document}

\title[Detecting Linguistic Bias in Government Documents Using Large language Models]{Detecting Linguistic Bias in Government Documents \\ Using Large language Models}
\author{Milena de Swart}
\email{milena.j.deswart@gmail.com}
\orcid{0009-0001-6997-7958}
\affiliation{%
    \institution{Ministerie van OCW}
    \city{the Hague}
    \country{the Netherlands}
}

\author{Floris Den Hengst}
\email{f.den.hengst@vu.nl}
\orcid{0000-0002-2092-9904}
\affiliation{%
  \institution{Vrije Universiteit Amsterdam}
  \city{Amsterdam}
  \country{the Netherlands}
}
\author{Jieying Chen}
\authornote{Corresponding author}
\email{j.y.chen@vu.nl}
\orcid{0000-0002-2497-645X}
\affiliation{%
  \institution{Vrije Universiteit Amsterdam}
  \city{Amsterdam}
  \country{the Netherlands}
}

\begin{abstract}
This paper addresses the critical need for detecting bias in government documents, an underexplored area with significant implications for governance. Existing methodologies often overlook the unique context and far-reaching impacts of governmental documents, potentially obscuring embedded biases that shape public policy and citizen-government interactions. To bridge this gap, we introduce the Dutch Government Data for Bias Detection (DGDB), a dataset sourced from the Dutch House of Representatives and annotated for bias by experts. We fine-tune several BERT-based models on this dataset and compare their performance with that of generative language models. Additionally, we conduct a comprehensive error analysis that includes explanations of the models’ predictions. Our findings demonstrate that fine-tuned models achieve strong performance and significantly outperform generative language models, indicating the effectiveness of DGDB~ for bias detection. This work underscores the importance of labeled datasets for bias detection in various languages and contributes to more equitable governance practices.
\end{abstract}




\maketitle

\section{Introduction}
\label{sec:introduction}
Linguistic bias emerges in various forms of communication, including online information and text, particularly when addressing categories of individuals or categorizing individuals~\cite{maass1999linguistic}. 
It is known to perpetuate stereotypes and increase discrimination, while also decreasing self-esteem, deteriorating mental and physical health, and reducing performance.

Linguistic bias can be defined as ``a systematic asymmetry in word choice as a function of the social category to which the target belongs''~\cite{beukeboom2017linguistic}. 
The words that refer to these social categories, as well as any terms that appear within such asymmetries in the context of these categories are referred to as \emph{biased terms}. Table~\ref{tab:example_sentences} lists some example sentences illustrating linguistic bias. The causes of linguistic bias lie with stereotypical expectations of the sender about the category of interest. Notably, linguistic bias may occur even when the sender is unaware of these expectations or when they do not intend to encode them in their communication. These characteristics make linguistic bias important to be mindful of, yet challenging to avoid in practice.

Researchers have therefore explored the use of automated tools for the detection of linguistic bias. Initially, these efforts mainly resulted in lists of terms associated with bias~\cite{nosek2002math}. Subsequent studies, for example, by \citet{recasens-etal-2013-linguistic}, have highlighted the role of linguistic context in identifying bias in text. Automated detection of bias has gained more prominence with the emergence of pre-trained generative language models (LMs) as these can handle large context windows \cite{blodgett2020language,garrido2021survey,gallegos2024bias}. 
The rise of generative LMs simultaneously exacerbates the problems associated with linguistic bias in source documents, as they may repeat or even amplify existing biases in these documents. 
Generative LMs therefore present both new opportunities to study bias in source documents while also imposing novel risks of perpetuating existing biases.

\begin{table*}[tbp]
\centering
\begin{tabular}{llc}
\toprule
Example sentence                                                        & Bias term group          & Bias \\
\midrule
\textbf{The handicapped} often have additional expenses                 & Prohibited        & $\checkmark$  \\
\textbf{Islam} considers the Quran to be literally written by Allah     & Conditionally biased       &               \\
\textbf{Islam} has therefore declared war on the Netherlands            & Conditionally biased       & $\checkmark$  \\
The \textbf{influx} of container shuttles by rail [...]                 & Context sensitive                          \\
2022 showed a large \textbf{influx} of refugees                         & Context sensitive           & $\checkmark$  \\
\bottomrule
\end{tabular}
\caption{Examples of different groups of bias terms with the term in \textbf{bold}.}
\Description{Examples of different groups of bias terms with the term in \textbf{bold}.}
\label{tab:example_sentences}
\end{table*}

\begin{table*}
\centering
\setlength{\tabcolsep}{5pt}
\begin{tabular}{lllr@{\hskip 0.04in}lllc}
\toprule
Dataset     &                                  & Task  & Size & Modality        & Language             & Text source         & Agreement \\ \midrule
DGDB        & \emph{ours}                              & B                 & 3,632         & sentences       & Dutch                         & Government    & $\kappa=.35$ \\ 
NPOV                & \citet{recasens-etal-2013-linguistic}    & B             & 230           & sentences       & English                       & Wikipedia      & $\kappa=.38\dag$ \\
DBWS                & \citet{hube2018detecting}                & B             & 685    & statements               & English                & Conservapedia & $\kappa=.35$ \\
AIBINA              & \citet{spinde2021automated}       & B             & 1,700         & sentences       & English                       & News           & $\kappa=.21$ \\
DiFair              & \citet{zakizadeh-etal-2023-difair} & GB             & 3,293       & sentence pairs  & English                       & Wikipedia               & $\kappa=.73\dag$ \\
GAP                 & \citet{webster-etal-2018-mind}     & GB             & 8,908       & pronouns–names & English                    & Wikipedia               & $\kappa=.74$ \\
DALC                & \citet{caselli2021dalc}           & H   & 8,156                   & messages        & Dutch                         & Twitter                 & $\kappa=.57$ \\
LiLaH               & \citet{hilte2023haters}            & H    & 10,732                & messages       & Multiple                      & Facebook                & $\alpha\leq.52$ \\
CrowS-pairs         & \citet{nangia2020crows}            & G              & 1,508       & sentence pairs  & English                       & Crowdsourced            & $\kappa=.46\dag$ \\
CrowS-pairs-F       & \citet{neveol2022french}           & G              & 1,679       & sentence pairs  & French                        & Crowdsourced            & $-$ \\
StereoSet           & \citet{nadeem2021stereoset}        & G              & 50,985      & sentences  & English                       & Wikipedia            & $-$ \\ 
HolisticBias        & \citet{smith-etal-2022-im}        & G              & 459,758      & sentences             & English                       & Crowdsourced            & $-$ \\
\bottomrule
\end{tabular}
\caption{Datasets for (gender) bias, hate speech and bias detection in generative models. Tasks: B=bias detection, GB=gender bias detection, H=hate speech detection, and G is bias detection in generative LMs. Agreement is expressed as Fleiss' $\kappa$ or Krippendorff's $\alpha$ when available, $\dag$ indicates values calculated from author-reported average and base rate agreement.}
\Description{datasets for (gender) bias, hate speech and bias detection in generative models. Tasks: B=bias detection, GB=gender bias detection, H=hate speech detection, and G is bias detection in generative LMs. Agreement is expressed as Fleiss' $\kappa$ or Krippendorff's $\alpha$ when available, $\dag$ indicates when values have been calculated from author-reported average and base rate agreement.}
\label{tab:related_work}
\setlength{\tabcolsep}{6pt}
\end{table*}

All efforts to automatically detect bias in linguistic texts benefit from the development of high-quality and realistic examples of biased text, and several datasets have been developed for this purpose. Most of these have focused on publicly available encyclopedia and news articles in English~\citep{recasens-etal-2013-linguistic,hube2018detecting,spinde2021automated},

but the creation of datasets that contain real-world examples of bias in various underrepresented languages and outside of the domain of news articles and social media remains challenging.

Existing methodologies focus on linguistic features of texts in general, often overlooking the distinctive context and broad societal impacts of government documents. This oversight can obscure how embedded biases in such texts not only perpetuate inequalities but also shape public policy and citizen-government interactions in profound ways. In this paper, we identify and address the pressing need for bias detection in  governmental documents, an area that remains underexplored despite its significant implications for governance and public administration, and therefore on the UN goals for sustainable development.
By focusing on government texts, this research aims to unveil the specific biases that influence legislative frameworks and policy directives, paving the way for more equitable governance practices. Our approach seeks to refine existing analytical tools to better understand and mitigate bias, ensuring that government communications uphold fairness and integrity, which is crucial to maintaining public trust and democratic ideals.

In this paper, we propose a methodology for creating bias detection datasets that generalize to various languages by utilizing publicly available government documents. 
We apply this methodology to create a bias detection dataset in the Dutch language and use the resulting dataset to train and evaluate the bias detection capabilities of various LMs in two regimes: an in-domain regime where sentences with biased terms are present during training and an out-of-domain regime where the models are evaluated on sentences with biased terms that were held out during training.

In our evaluations, we find that trained models significantly outperform untrained models on this task, including pre-trained generative LMs. 
This implies that DGDB~ is sufficiently rich to learn how to address the challenging task of detecting bias based on the linguistic context. It thereby not only contributes to the academic discourse on linguistic bias in general and in government texts specifically, but also serves as a vital resource for policymakers and educators aiming to create more equitable societies.

Finally, our findings highlight the importance of specialized datasets for bias detection, especially in the presence of large generative LMs. The presented models can be used to detect bias in Dutch text in practice, the dataset can be used to further study how to protect underrepresented communities under Dutch governance, while the presented approach to collect this dataset may inspire the development of additional bias detection datasets in other underrepresented languages. The methodologies introduced for creating a bias detection dataset and for training bias detection models may thus have utility across linguistic boundaries in fostering fair and unbiased communication.

\section{Related Work}
\label{sec:related_work}

We examine related datasets and models for bias detection and compare them to our contributions. Table~\ref{tab:related_work} summarizes various efforts, detailing the specific tasks addressed, data modalities, and dataset sizes.

Among the tasks explored in prior work, some datasets are dedicated specifically to gender bias detection and hate speech detection. These can be considered subtasks of the broader bias detection field: gender bias focuses on a particular target group, while hate speech is typically intentional. In contrast, general bias detection encompasses various target groups and may involve both intentional and unintentional biases. Table~\ref{tab:example_sentences} presents examples that highlight the differences between these concepts. Additionally, Table~\ref{tab:related_work} includes datasets designed to study the existence of biases and stereotypes in generative LMs. 
While these datasets contain labelled sentences, they often consist of (masked) sentences used to test whether generative LMs prefer certain tokens, thereby measuring the inherent bias in these models.

Existing datasets targeting bias detection are generally of moderate size and are derived from news articles and encyclopedias in English. There are two main reasons for this. First, these sources are extensively used by the computational linguistics and natural language processing communities and are therefore readily accessible. Second, these sources typically strive to exclude linguistic biases from their content to some degree, making the detection of bias both technically challenging and practically relevant. The texts in our dataset are also assumed to be written without intentional bias but are sourced from government documents rather than the commonly used sources within the community.

Several models have been proposed for the detection of linguistic bias. Some focus on detecting bias at the word level~\cite{recasens-etal-2013-linguistic,hube2018detecting,spinde2021automated}, while others emphasize the importance of context in bias detection~\cite{kuang2016semantic,recasens-etal-2013-linguistic,bartl-etal-2020-unmasking,zhang-etal-2020-demographics}. While word-level bias detection aids in understanding specific linguistic markers of bias, utilizing context is crucial for detecting subtle forms of bias and is generally considered more suitable for bias detection tasks. Therefore, we focus on models that take linguistic context into account when detecting bias within DGDB.

\section{DGDB: a Bias Detection Dataset from Government Documents}
\label{sec:dataset}
\begin{figure}[t]
\centering
  \includegraphics[width=\columnwidth]{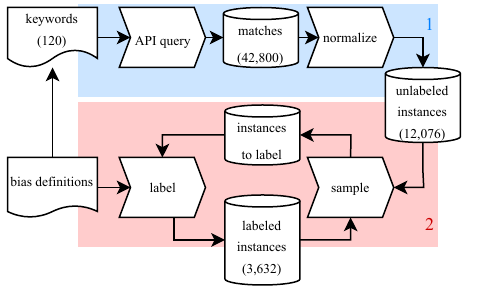}
  \caption{Dataset Creation Process. The initial automated step (highlighted in blue) utilized bias-related keywords from established definition sources to query a web API and retrieve relevant documents. Subsequently, these normalized documents were iteratively sampled and annotated by bias detection experts in the second step (highlighted in red).}
  \Description{dataset creation process.}
  \label{fig:process}
\end{figure}

This section details the collection, annotation, and compilation of publicly available government documents into the Dutch Government Data for Bias Detection (DGDB) dataset. The dataset was created through a multistep process, as illustrated in Figure~\ref{fig:process}. 

Initially, we defined various types of linguistic bias to develop a comprehensive list of associated keywords. These keywords were utilized to query publicly available government documents from the Dutch House of Representatives (Tweede Kamer). From these documents, sentences containing bias-related terms and their contextual information were extracted, mined, and normalized. This process resulted in a collection of 42,800 matches spanning over 238,000 pages, which were then deduplicated to ensure data quality.

Subsequently, 12,076 sentences identified as at risk of bias were sampled and annotated by a team of 14 expert annotators specializing in discrimination, racism, and bias detection. Through iterative sampling and expert labeling, a final dataset of 3,632 sentences with annotations was compiled. This meticulous process ensures that the dataset accurately reflects various forms of bias as defined in our guidelines. The dataset\footnote{\url{https://huggingface.co/datasets/milenamileentje/Dutch-Government-Data-for-Bias-detection/}}, models\footnote{\url{https://huggingface.co/milenamileentje/BiasBERTje}} and code\footnote{\url{https://github.com/milenadeswart/Datalab_PDR/tree/Milena}} are publicly available.

\subsection{Bias Definition}
\label{sec:definition}
A comprehensive keyword list was created by integrating established sources that advocate for social inclusion and address racism, discrimination, and bias within the fields of cultural policy and art curation. 
These sources were developed in collaboration with advisory panels representing diverse perspectives and backgrounds~\cite{modest2018woorden,samuel2022waarden}.
The list includes examples and detailed clarifications to distinguish between biased and unbiased language usage.

The terms used in this work are grouped into one of nine categories to express the kind of bias they signal: \texttt{algemeen} (general),
\texttt{beperkingen} (disabilities), \texttt{cultuur} (culture), \texttt{geloof} (religion), \texttt{gender} (gender),
\texttt{kolonialisme} (colonialism), \texttt{migratie} (migration), \texttt{onderwijs} (education), and \texttt{seksualiteit}
(sexuality). 
Among these categories, \texttt{kolonialisme} is the largest with 28 terms, and \texttt{cultuur} contains the fewest terms, with only three. 
In most cases, the context in which these terms are used is essential for identifying the presence of bias.

To streamline the annotation process, the terms were systematically organized into three distinct groups:
\begin{description}
\item[Prohibited] These terms should only be used in direct quotations or to explain why they are harmful; otherwise, they should be avoided. In many cases, alternatives are available, such as replacing \texttt{gehandicapte} (handicapped person) with \texttt{persoon met beperkingen} (person with disabilities).

\item[Conditionally biased] Terms that exhibit bias depending on the context. 
This includes (near-)homonyms, such as the noun \texttt{Christen} (Christian) referring to `individual(s) subscribing to a particular religion' and the adjective \texttt{Christelijk} (Christian), which may signify a specific group and could indicate bias based on its usage.

\item[Context sensitive] Terms where the potential for bias depends heavily on the context, e.g. which are neutral in isolation but can become problematic in certain contexts, such as \texttt{achtergrond} (origin), \texttt{ouders} (parents), \texttt{man} (man), etc.

\end{description}
These categories were only used to facilitate the annotation process, and are not part of the bias detection task or DGDB.
 The comprehensive list of terms can be found in the Appendix~\ref{app:bias_terms}.

\begin{definition}
A sentence is classified to be \texttt{biased}
if it satisfies any of the following conditions:

\begin{description}
\item[1] The sentence contains a term from the \textit{`prohibited’} category, indicating that the term is universally inappropriate, \textbf{OR}
\item[2a] The sentence exhibits bias, such as stereotyping, exclusion, power imbalance, or prejudice—whether implicit or explicit, \textbf{AND}
\item[2b] That bias is directed toward a specific group.
\end{description}
\end{definition}
Table~\ref{tab:example_sentences} provides examples of linguistically biased sentences, alongside comparisons with hate speech and gender bias detection.

\subsection{Data Collection}
\label{sec:collection}

After compiling a list of 120 keywords, Dutch government data from the Dutch House of Representatives (Tweede Kamer) were queried to identify documents containing one or more matches (step 1, highlighted in blue in Figure~\ref{fig:process}). To ensure consistency, relevance, and alignment with contemporary discourse, we focused on documents from 2010 onward, restricting the collection to those less than 15 years old.

A total of 42,800 matches were identified, spanning over 238,000 pages. These documents were scraped, deduplicated, and normalized, resulting in 12,076 sentences flagged as potentially biased. The textual context of each match was preserved throughout the process. Specifically, all documents were first collected and split into sentences, with capitalization removed for uniformity.

The collected documents primarily consist of formal government communications, including ministerial notes (\texttt{Nota}), explanatory memoranda (\texttt{Memories van Toelichting}), ministerial reports (\texttt{Rapporten}), and attachments to these documents.

\subsection{Data Annotation}
\label{sec:annotation}
Data annotation (step 2 marked in red in Figure~\ref{fig:process}) was conducted by a team of 14 experts from government employees.
These annotators were all native speakers, familiar with government documents, and specialized in areas such as discrimination, racism, and bias. Participation in the annotation process was voluntary.

The annotators followed to the set of guidelines for identifying bias in sentences, as outlined in Section~\ref{sec:definition}. 
These guidelines, agreed upon by the annotation team, were adapted from those of~\citet{nangia2020crows}. The definitions were designed to capture a broad spectrum of bias, ranging from subtle and possibly unintended language variations to more explicit forms, such as hate speech~\cite{fortuna2018survey}. For example, sentences containing positive prejudices (e.g., "women are more emotionally mature") were labeled as \texttt{biased} if they implied negative prejudices against other groups (e.g., "men are less or not emotionally mature").

The annotation process was divided into three separate annotation sessions.
Each session began with discussions to align definitions and refine processes, followed by a training session to introduce the project context and clarify annotation guidelines. A playtest labeling session with an initial set of seven annotators was conducted to test and refine the instructions. Based on feedback, additional examples were incorporated into the guidelines to improve clarity and reduce ambiguity. Annotators participated in sessions based on availability, ensuring that each session included experts on bias, discrimination, and racism.

The annotators were asked to assign a label 0 (\texttt{not biased}),
a label 1 (\texttt{biased}), a label 2 (\texttt{unsure}), or a label 3 (to be \texttt{excluded} from the dataset).
Sentences labeled as 2 (\texttt{unsure}) were later re-annotated to ensure precision. Annotators were encouraged to discuss challenging cases collaboratively, fostering consistency and alignment across the labeled dataset.

The annotation sessions proceeded as follows.
 In the first session, seven annotators labeled a random sample of 1,800 instances.
 In the second session, eight annotators, including three from the first session, labeled 1,547 randomly selected instances.
 The final session focused on increasing the diversity of terms, annotating 1,007 instances containing terms not yet present in the labeled dataset.

Annotators reported that the task was challenging, particularly due to the nuanced nature of the definitions and due to unfamiliarity with the task. However, refinements to the guidelines and the inclusion of additional examples after playtesting made the process more manageable.

The final dataset includes 3,747 sentences labeled as \texttt{biased} or \texttt{not biased}. Inter-annotator agreement (IAA) was evaluated using Fleiss’ $\kappa$, based on 255 annotations from three of the most experienced annotators, with an expert resolving uncertain cases~\cite{fleiss1973equivalence}. The IAA was calculated as $\kappa=0.35$. Although this value is relatively low compared to other tasks, it aligns with similar datasets in this domain. Disagreements often involved nuanced cases, such as the sentence, ``The Tax Authority is working with affected parents [...]," where the term ``parents" could be interpreted as bias by excluding ``caregivers" as a broader category.

\subsection{Characteristics}
\label{sec:dataset_details}
\begin{table}

\centering
\begin{tabular}{lr}
\toprule
Category            & Occurence    \\ \midrule
Colonialism         & 220           \\
Culture             & 43            \\
Disabilities        & 200           \\
Education           & 1,078         \\
Gender              & 630           \\
General             & 803           \\
Migration           & 1,501         \\
Religion            & 140           \\
Sexual orientation  & 499           \\
\bottomrule
\end{tabular}

\caption{Occurrences of bias terms by category.}
\label{tab:category_distribution}
\end{table}
The DGDB~dataset contains 3,747 annotated sentences with 948 instances (25.3\%) labeled as \texttt{biased}, making it a reasonably imbalanced dataset. Of the 120 terms in the bias keyword list, 48 are present in the final dataset. The most occurring term is present 560 times, and the lowest term occurrence is one.
Table~\ref{tab:category_distribution} shows the distribution of the bias categories.
Data were split into train, test, and validation splits of proportions 0.6, 0.2, 0.2 of the entire dataset, respectively. The annotated dataset DGDB~is publicly available and published under a permissive license.

\section{Evaluation Models for Bias Detection}
\label{sec:models}
\begin{table*}[t]
\centering
  \begin{tabular}{llccc}
  \hline
  Model             &                                           & Language              & Train Task                    & Training data  \\\hline
  BERTje            & \citet{devries2019bertje}                 & Dutch                 & Bias detection                & DGDB              \\
  RobBERT           & \citet{delobelle-etal-2020-robbert}       & Dutch                 & Bias detection                & DGDB              \\
  SentenceBERT      & \citet{reimers2019sentence}               & Multiple              & Bias detection                & DGDB              \\
  DALC-BERT         & \citet{caselli2021dalc}                   & Dutch                 & Hate speech detection         & DALC              \\
  GPT-3.5           &                                           & Multiple              & Generic                       & $-$               \\
  GPT-4o mini       &                                           & Multiple              & Generic                       & $-$               \\
  \hline
  \end{tabular}
  \caption{Overview of models.}
  \label{tab:models}
\end{table*}
This section outlines the development and evaluation of various models for bias detection using the compiled Dutch Government Data for Bias Detection (DGDB) dataset. 

\subsection{Experimental Setup}
\label{sec:experimental_setup}
Our experiments are designed to investigate the utility of DGDB~in training and evaluating bias detection models, focusing on predictive performance, generalization to unseen terms, and strategies to address the dataset's imbalanced nature. We address the following research questions:
\begin{description}
    \item[RQ1] Which model most effectively detects bias?
    \item[RQ2] How well do models generalize to biased terms not seen during training?
    \item[RQ3] What are the effects of resampling techniques on model performance?
    \item[RQ4] How does contextual information influence the detection of biased terms?
\end{description}

To address RQ1, we compare the performance of various BERT-based models fine-tuned on DGDB~against generative LMs, as detailed in Section~\ref{sec:training}. 

For RQ2, we evaluate model generalization by training in two distinct regimes:
\begin{itemize}
    \item \textbf{In-domain regime}: Models are trained on the entire dataset.
    \item \textbf{Out-of-domain regime}: Models are trained with certain rare bias terms excluded. Specifically, terms appearing ten times or fewer in the dataset are omitted, resulting in the exclusion of 12 terms and 69 instances. This approach allows us to assess how well models handle bias terms that were not present during training.
\end{itemize}
We then evaluate these models on a test set containing the excluded rare terms to determine their ability to generalize to unseen bias terms.
Given that DGDB~is moderately imbalanced, with 25.3\% of instances labeled as \texttt{biased}, we explore the impact of three resampling strategies to mitigate class imbalance for RQ3:
\begin{itemize}
    \item \textbf{Oversampling}: Duplicate \texttt{biased} instances to increase their proportion to 38.6\% in a training set of 2,648 instances.
    \item \textbf{Undersampling}: Randomly sample \texttt{unbiased} instances to achieve a class ratio of 31.0\% \texttt{biased} in a training set of 1,649 instances.
    \item \textbf{Balanced Resampling}: Combine oversampling and undersampling to create a balanced training set of 2,137 instances with an equal distribution of \texttt{biased} and \texttt{not biased} labels (50.0\% each).
\end{itemize}

Finally, RQ4 is addressed through a quantitative error analysis and a qualitative investigation using the LIME framework~\cite{ribeiro2016should}. This analysis explores how contextual information influences the model's ability to detect biased terms accurately.

\subsection{Model and Prompting Details}
\label{sec:training}

We selected models based on insights from prior research (see Section~\ref{sec:related_work}). Table~\ref{tab:models} provides an overview of all models employed in this study. Primarily, we utilized various BERT models specifically tailored for the Dutch language, chosen for their demonstrated effectiveness in related tasks such as hate speech detection~\cite{caselli2021dalc} and their computational efficiency. Specifically, three BERT-based models were fine-tuned on the DGDB dataset. Additionally, we incorporated a model specifically trained for hate speech detection in Dutch, referred to as ``DALC-BERT''. To ensure a comprehensive comparison, we also evaluated three multilingual generative LMs.

For fine-tuning the BERT-based models, we adhered to best practices and conducted preliminary experiments as outlined by Sun et al.~\cite{sun2019fine}. The following hyperparameters were used:
\begin{itemize}
    \item Optimizer: ADAM with $\beta_1 = 0.9$ and $\beta_2 = 0.999$
    \item Batch size: 8
    \item Dropout rate: 0.1
    \item Learning rate: $2 \times 10^{-5}$
    \item Number of epochs: 4 (trained until convergence)
\end{itemize}

In the experiments involving generative LMs, we developed zero-shot prompts that closely aligned with our annotation process, maintaining consistent definitions, expected outputs, and criteria. During the initial prompt development phase, we evaluated both English and Dutch prompts, ultimately selecting the Dutch prompts due to their superior performance. We also experimented with chain-of-thought prompting. 
However, this approach did not enhance labeling accuracy and resulted in inconsistent output formatting.
The final prompts used for generative LMs, in both Dutch and English, are presented in Figure~\ref{fig:prompt}.

\begin{figure*}[ht]
\centering
\small
\begin{minipage}[t]{0.48\textwidth}
\fbox{
\parbox{0.9\linewidth}{
\textit{``Je bent een expert op het gebied van bias, stereotypering en discriminatie. Je taak is om zinnen te classificeren volgens de volgende definitie: een zin bevat bias als de zin een stereotype, uitsluiting, machtsverhoudingen of vooroordelen bevat, die impliciet of expliciet kunnen zijn, en die gaan over een af te bakenen groep personen, OF als een zin een woord bevat wat moet worden vermeden omdat het bias bevat. Geef de volgende zin een label, antwoord alleen met 0 of 1. Geef de zin het label 0 voor niet-biased en 1 voor biased. De zin is: [item]."}
}
}
\end{minipage}
\hfill
\begin{minipage}[t]{0.48\textwidth}
\fbox{
\parbox{0.9\linewidth}{
\textit{``You are an expert in bias, stereotyping, and discrimination. Your task is to classify sentences according to the following definition: a sentence contains bias if it includes stereotypes, exclusion, power dynamics, or prejudices—which can be implicit or explicit—about a specific group of people, OR if the sentence contains a word that should be avoided because it is biased. Label the following sentence by answering only with 0 or 1. Assign the label 0 for not biased and 1 for biased. The sentence is: [item]."}
}
}
\end{minipage}
\caption{Prompts for generative LMs in Dutch (left) and their English translation (right).}
\label{fig:prompt}
\end{figure*}

\subsection{Results}
\label{sec:results}

\begin{figure*}[t]
    \centering
    \begin{tikzpicture}
    \begin{axis}[
    ybar,
    width=\textwidth,
    height=150,
    ymin=0,
    ymax=1.24,
    enlarge x limits=0.02,
    symbolic x coords={newcomer,slavery history,Islam,minorities,gap,gender,boy,girl,transgender,Jewish,stream down,headscarf,parents,migrants,other language,poverty,flow,sexual,Christian,tradition,manpower,disabled,power,current,Arabic,heterosexual,homosexual,highly educated,lowly educated,privilege,racism,slave,repeating student},
    xtick=data,
    bar width=0.2cm,
    xticklabel style={rotate=45,anchor=east}, 
    nodes near coords,
    nodes near coords align={vertical},
    nodes near coords style={font=\small, rotate=90, anchor=west}
    ]
    \addplot[ybar, fill=blue, color=blue] coordinates {(newcomer,0.0) (slavery history,0.3333333333333333) (Islam,0.5) (minorities,0.5) (gap,0.6) (gender,0.6181818181818182) (boy,0.6666666666666666) (girl,0.6666666666666666) (transgender,0.6666666666666666) (Jewish,0.7142857142857143) (stream down,0.75) (headscarf,0.75) (parents,0.7647058823529411) (migrants,0.78125) (other language,0.8) (poverty,0.8780487804878049) (flow,0.8878504672897196) (sexual,0.8904109589041096) (Christian,0.9) (tradition,0.9166666666666666) (manpower,0.9333333333333333) (disabled,0.9354838709677419) (power,0.958904109589041) (current,0.9893617021276596) (Arabic,1.0) (heterosexual,1.0) (homosexual,1.0) (highly educated,1.0) (lowly educated,1.0) (privilege,1.0) (racism,1.0) (slave,1.0) (repeating student,1.0)};
    \end{axis}
    \end{tikzpicture}  
  \caption{Accuracy of bias detection over terms, relative to term occurrence (in-domain).}
  \Description{Accuracy of bias detection over terms, relative to term occurrence (in-domain).}
  \label{fig:error_indom}
\end{figure*}

\begin{figure}[t]
  \begin{tikzpicture}
    \begin{axis}[
    ybar,
    ymin=0,
    symbolic x coords={underachiever,eastern bloc,slaves,refugee crisis,in love,female teacher,to flow up,stay-at-home,fobia,fair (of skin),indigenous,intermediate position},
    xtick=data,
    xticklabel style={rotate=45,anchor=east}, 
    nodes near coords,
    nodes near coords align={vertical},
    nodes near coords style={font=\small, rotate=90, anchor=west},
    height=5cm,
    width=\columnwidth,
    ]
    \addplot[ybar, fill=blue, color=blue] coordinates {(underachiever,0.0) (eastern bloc,0.0) (slaves,0.0) (refugee crisis,0.125) (in love,0.25) (female teacher,0.4444444444444444) (to flow up,0.4444444444444444) (stay-at-home,0.5) (fobia,0.8) (fair (of skin),1.0) (indigenous,1.0) (intermediate position,1.0) 
};
    \end{axis}
    \end{tikzpicture} 
  
  \caption{Accuracy of bias detection over terms, relative to term occurence (out-of-domain).}
  \Description{Accuracy of bias detection over terms, relative to term occurence (out-of-domain).}
  \label{fig:error_outdom}
\end{figure}

Table~\ref{tab:main_results} presents the F1 scores~\footnote{As this is a binary task, micro-F1 equals macro-F1} for the in-domain and out-of-domain regimes across models on the original dataset, highlighting the best-performing sampling strategy for each model. The results demonstrate that reasonable performance can be achieved in the challenging bias detection task by training BERT-based models on DGDB~in the in-domain regime (RQ1). The significant performance improvement from using a bias-specific dataset underscores the continued necessity of labeled datasets for effective bias detection, even with the advancements in generative LMs. However, in the out-of-domain regime, performance declines for these models, though they still outperform the random baseline (RQ2). Notably, generative LMs and BERT-based models show comparable performance in this regime. These findings indicate that while bias detection is feasible, the models’ generalization remains a challenge, necessitating retraining as new bias-associated terms emerge.
\begin{table}[tbp]
\centering
\begin{tabular}{lcc}
\toprule 
Model                   & In-domain         & Out-of-domain     \\\midrule
BERTje                  & \textbf{.812}    & .554             \\
RobBERT                 & .811             & .499             \\
SentenceBERT            & .785             & .390             \\
DALC-BERT               & .792             & .531             \\ \hline
GPT-3.5                 & -                 & \textbf{.559}    \\
GPT-4o mini             & -                 & .550             \\
\bottomrule
\end{tabular}
\caption{F1 test scores (per model best in undersampling regime), best in \textbf{bold}.}
\label{tab:main_results}
\end{table}

Table~\ref{tab:sampling_results} details the performance of various resampling strategies for BERTje. While the overall impact of sampling is limited, undersampling consistently produces the best results (RQ3). This is particularly noteworthy as it suggests that the advantages of a more balanced dataset outweigh the potential drawbacks of a reduced training set size. These findings highlight the importance of smaller, high-quality datasets over larger, low-quality ones for bias detection tasks. Future efforts should prioritize creating well-curated datasets that comprehensively represent linguistic bias to improve model performance and generalization.

\begin{table}[t]
\centering
\begin{tabular}{lcc}
\toprule 
Strategy                & In-domain         & Out-of-domain     \\\midrule
No sampling             & .793              & .435             \\
Undersampling           & \textbf{.812}     & \textbf{.554}     \\
Oversampling            & .783              & .456              \\
Balanced       & .791              & .435              \\
\bottomrule
\end{tabular}
\caption{F1 test scores for BERTje, best in \textbf{bold}.}
\label{tab:sampling_results}
\end{table}

\subsection{Error Analysis}
\label{sec:error_analysis}
\begin{figure}[t]
  \includegraphics[width=0.7\columnwidth]{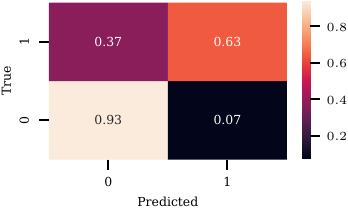}
  \caption{Confusion matrix for BERTje w/ undersampling.}
  \Description{Confusion matrix for BERTje w/ undersampling.}
  \label{fig:confusion_matrix}
\end{figure}
To address Research Question 4 (RQ4), we conduct an error analysis to understand the types of mistakes made by the model. An initial step in evaluating a model's performance and behavior involves examining its error patterns. Figure~\ref{fig:confusion_matrix} presents the confusion matrix for the BERTje model trained with undersampling. The model demonstrates high accuracy on \texttt{unbiased} instances, correctly classifying 93\% of these cases, compared to 63\% accuracy on \texttt{biased} instances. This discrepancy is primarily attributed to the remaining class imbalance despite resampling efforts.

We proceed with a more granular quantitative error analysis by investigating the accuracy per bias term. Figure~\ref{fig:error_indom} lists the accuracy for each term in the best-performing model within the in-domain regime. The results reveal significant variability in model performance across different terms. For example, the term \texttt{nieuwkomer} (newcomer) was incorrectly classified in all instances within the test set. In contrast, several other terms were accurately classified in every occurrence. This latter group includes terms from the \texttt{prohibited} category and those whose bias is context-dependent, such as \texttt{stroom} (influx). These findings imply the necessity of high-quality, comprehensive training data to reliably predict bias across diverse terms and contexts.

Figure~\ref{fig:error_outdom} illustrates the per-term accuracy of the best-performing model in the out-of-domain regime. The results indicate that the model struggles to accurately classify terms that were not present during training. Similar to the in-domain results, there is substantial variability in performance across different terms. This suggests that while contextual information aids in bias detection, it is insufficient on its own. The models require diverse and extensive training datasets to effectively generalize and detect bias in novel terms.

\begin{figure}[t]

  \hfill\includegraphics[width=0.84\columnwidth]{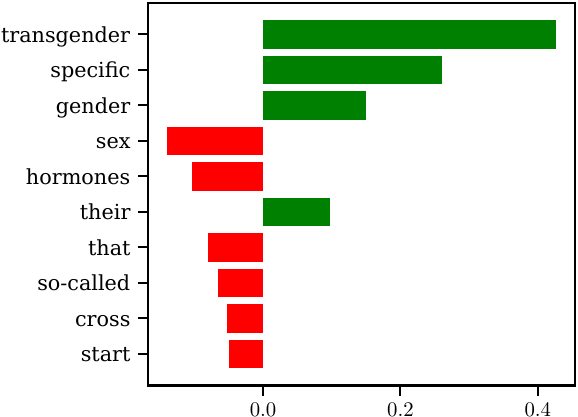}

  \hfill\includegraphics[width=\columnwidth]{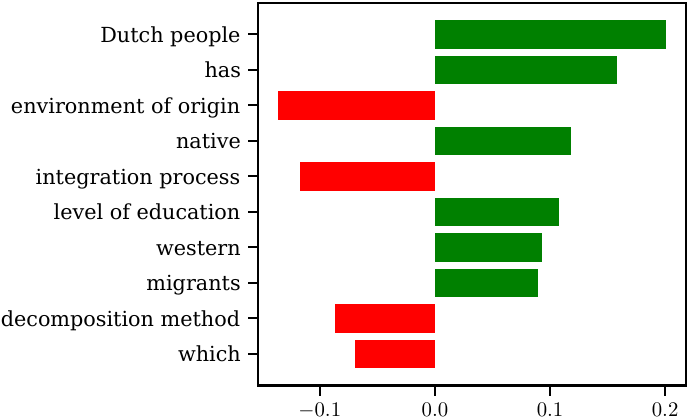}
  
  \hfill\includegraphics[width=0.96\columnwidth]{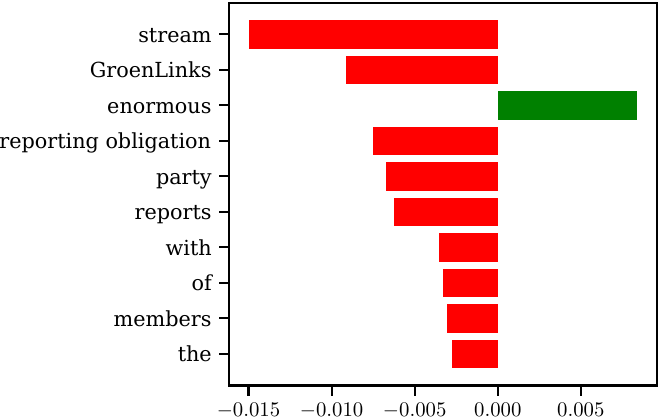}

  \caption{Model explanation using LIME for two positive (top, center) and one negative input (bottom).}
  \Description{Model explanation using LIME.}
  \label{fig:LIME_example}
\end{figure}

We conclude our error analysis by examining examples of model predictions using the LIME framework. Figure~\ref{fig:LIME_example} illustrates one such analysis. Our findings reveal that the keyword terms used during the search significantly contribute to classifying instances as \texttt{biased}, reinforcing the critical role of labeled datasets in bias detection.
Additionally, terms related to (ethnic) background, gender, and sex often strongly influence predictions toward \texttt{biased} classifications. In contrast, terms associated with neutrality or objectivity, such as ``environment of origin”, ``integration process”, ``decomposition method”, and ``reports”, tend to influence against \texttt{biased} classifications.

\section{Implications for Sustainable Development}
\label{sec:implications}

The approach, dataset, and models presented in this work align with several UN Sustainable Development Goals. Specifically, they support Goal 11 (Make cities and human settlements inclusive, safe, resilient and sustainable) and Goal 16 (Promote peaceful and inclusive societies, provide access to justice for all, and build effective, accountable institutions). By enabling governments to monitor and mitigate biased language in their communications, we also contribute to Goal 5 (Achieve gender equality and empower all women and girls) and Goal 10 (Reduce inequality within and among countries) by providing tools and methodologies for bias detection.

By enabling governments to monitor and address bias in their communications, this work provides tools and methodologies for detecting bias, fostering inclusivity, and improving policy development. Policymakers can identify and revise language that unintentionally marginalizes certain groups or perpetuates stereotypes, leading to more equitable legislation and communication.

\section{Limitations}
While this study provides valuable insights, it is important to acknowledge certain limitations that offer avenues for future research. The data collection methodology presented here is adaptable to other languages.
However, it requires the availability of government data for querying, which may vary across different regions. 
Additionally, our approach relies on a predefined seed list of bias terms, which presents two key considerations. 
First, the seed list may not be exhaustive, potentially omitting some forms of bias from the resulting dataset. Nonetheless, this challenge persists even without a seed list, as perceptions and manifestations of bias evolve over time. Second, the development and consensus on such a word list are essential for consistent application. 
We consider this limitation to be relatively minor, as the seed list used can serve as a foundation for creating comprehensive bias term lists in other languages through (automated) translation. Overall, these limitations highlight opportunities to enhance the methodology and extend its applicability to diverse linguistic and cultural contexts, thereby strengthening the robustness and inclusivity of bias detection efforts.

\section{Ethical Considerations}

In addressing the critical need for bias detection in government texts, we emphasize several ethical considerations essential for the responsible application of AI in governance.

\textbf{Data Privacy and Security.} First and foremost, we ensure strict adherence to data privacy standards, particularly in handling sensitive government documents. This is achieved through rigorous anonymization procedures and robust data storage protocols, safeguarding the confidentiality and integrity of the data.

\textbf{Bias in Dataset Creation.} We critically examine potential biases within our newly introduced Dutch Government Data for Bias Detection dataset, addressing both selection and annotation biases to ensure a fair and representative dataset. By meticulously curating the data, we aim to minimize inherent biases that could skew the results and perpetuate unfair representations.

\textbf{Transparency and Explainability.} Transparency in our methodology is paramount. We provide a detailed account of the processes involved in training and fine-tuning our BERT-based models, supported by a comprehensive explainability framework. This approach ensures that model decisions are interpretable and accountable, aligning with ethical guidelines such as the EU’s Ethics Guidelines for Trustworthy AI~\cite{EU_Trustworthy_AI}. By making our methods and model behaviors transparent, we foster trust and enable scrutiny from the broader community.

\textbf{Accountability and Fairness.} Our models are designed to be accountable, with mechanisms in place to address and mitigate any unintended biases that may arise. We continuously evaluate and refine our models to uphold fairness and prevent the reinforcement of existing prejudices within governmental operations.

\textbf{Impact on Public Policy.} We carefully consider the potential impact of our findings on public policy. Enhanced bias detection capabilities can inform more equitable policy-making and governance, promoting fairness and reducing discrimination in public administration. Our work aims to support policymakers in creating inclusive and unbiased policies that reflect diverse societal needs.

\textbf{Future Directions.} Future research will continue to refine these ethical approaches, ensuring that advancements in AI capabilities are matched by robust ethical considerations. We are committed to fostering fairness, integrity, and accountability in governmental operations through the responsible use of AI technologies.

\section{Conclusion}
\label{sec:conclusion}

In this study, we introduce the DGDB, a novel dataset comprising 3,747 annotated sentences sourced from the Dutch House of Representatives. This dataset was utilized to train and evaluate four variations of BERT-based models, demonstrating its effectiveness in identifying linguistic bias and providing insights into language variations associated with bias.

Our results indicate that government documents are a viable source for bias detection data, underscoring the complexity of the bias detection task. The fine-tuned BERT-based models achieved strong performance in the in-domain regime, highlighting the importance of specialized, labeled datasets for bias detection, especially when leveraging LMs. Although performance declined in the out-of-domain regime, models still surpassed the random baseline, suggesting that while bias detection is achievable, enhancing generalization remains a challenge. Additionally, our investigation into resampling strategies revealed that undersampling the majority class improves model performance, emphasizing the value of balanced, high-quality datasets over larger, imbalanced ones.

The DGDB~dataset and the trained models presented in this work hold potential for reducing linguistic bias in governmental communications by serving as tools for bias monitoring and correction. By implementing such bias detection mechanisms, government bodies can promote more equitable and unbiased policy-making and governance.

Future work should focus on expanding DGDB~to encompass a broader range of languages and cultural contexts, thereby enhancing the generalizability of bias detection models. Additionally, incorporating more labels and integrating uncertainty quantification frameworks can further improve model reliability and user trust. Combining these models with interactive classification systems can also facilitate real-time bias detection and mitigation in governmental documents~\cite{angelopoulosconformal,hengst2024conformal}.

\begin{acks}
The authors gratefully acknowledge the support provided by the Ministry of Education, Culture and Science of the Netherlands
and the Ministry of the Interior and Kingdom Relations of the Netherlands
Special thanks are extended to Coen Eisma, Sybren Spit, and Denise Heiligers from the Directie Kennis OCW for their invaluable contributions. The authors also sincerely thank the contributors to the annotation process: Alex Mulder, Amy Chen, Audrey Wei-Yun, Cheyenne Ramada, Coen Eisma, Dennis van Gessel, Irene van der Vossen, Jeroen Vugs, Michel Pan, Michelle Claus, Tess Rijk, Tjebbe Monasch, Tristan Berlijn, and Thijs Noordzij. This research was also partially funded by the Hybrid Intelligence Centre, a 10-year programme supported by the Dutch Ministry of Education, Culture and Science through the Netherlands Organisation for Scientific Research (NWO), under grant number 024.004.022.
\end{acks}

\bibliographystyle{ACM-Reference-Format}
\balance
\bibliography{bibliography}

\appendix
\onecolumn

\section{List of bias terms}
\label{app:bias_terms}
Table~\ref{tab:searchwordlist} shows all keywords and their categories, sourced from various Dutch bias definition sources~\cite{samuel2022waarden,modest2018woorden}.
\begin{table*}[h!]
  \small
  \begin{tabularx}{\textwidth}{l |X}
    \toprule
    \textbf{Category} & \textbf{Keywords} \\
    \midrule
    Algemeen & Andere achtergrond, armoede, fobie, grensoverschrijdend gedrag, macht, minderheden \\ 
    
    Beperkingen & handicap, dwerg, begeleider, doventolk, gebarentolk, rolstoeler, rolstoelrijder, rolstoelgebonden, gehandicapt \\ 
    
    Cultuur & bi-cultureel, traditie, tussenpositie \\
    
    Geloof & christen, hoofddoek, Islamiet, Mohammedaan, islam, joods \\
    
    Gender & hermafrodiet, dames en heren, hij of zij, hij/zij, geslacht, jongen, kloof, man, meisje, non-binair, vrouw, transseksueel, travestiet, transgender \\
    
    Kolonialisme & etnisch, bruin, donker, meerbloed, dubbelbloed, halfbloed, anderstalig, indiaan, medicijnman, ontdekken, ontdekking, blank, caribisch gebied, kaukasisch, gouden eeuw, inheems, exotisch, gekleurd, primitief, ras, page, bediende, Eskimo, slaaf, slaven, slavernijverleden, zwart \\
    
    Migratie & Berber, Turk, autochtoon, allochtoon, nieuwe Nederlander, bicultureel, west, zigeuner, migranten, nieuwkomer, ontwikkelingslanden, lagelonenlanden, derde wereld, oostblok, stromen, stroom, westers, vluchtelingencrisis \\
    
    Onderwijs & achterstandsleerling, achterstandsscholen, achterstandsschool, achterstandsscore, afstromen, arabisch, excellente school, hoogopgeleid, juf, laagopgeleid, mavo, mbo-cursist, mbo-deelnemer, mbo-leerling, opstromen, ouders, passend onderwijs, plusklassen, praktisch geschoold, speciaal onderwijs, theoretisch geschoold, thuiszitter, witte school, zittenblijven, zwarte school, probleemkinderen, probleemwijk, probleembuurt, laag opgeleid, lager opgeleid \\
    
    Seksualiteit & hetero, homo, queer, seksueel, verliefd \\
    
    \bottomrule
  \end{tabularx}
  \centering
    \caption{Categorized Bias Keywords for Detection}
      \label{tab:searchwordlist}
\end{table*}
\FloatBarrier 
\clearpage

\section{Translated Versions of Figures}
This appendix includes various versions of the figures from the main body, along with their translated text to provide clarity and accessibility for a wider audience.

\begin{figure*}[b]
    \centering
    \begin{tikzpicture}
    \begin{axis}[
    ybar,
    width=\textwidth,
    height=200,
    ymin=0,
    ymax=1.15,
    symbolic x coords={afstromen,anderstalig,arabisch,armoede,christen,gehandicapt,geslacht,hetero,homo,hoofddoek,hoogopgeleid,islam,jongen,joods,kloof,laagopgeleid,macht,mankracht,meisje,migranten,minderheden,nieuwkomer,ouders,privilege,racisme,seksueel,slaaf,slavernijverleden,stromen,stroom,traditie,transgender,zittenblijven},
    xtick=data,
    bar width=0.2cm,
    xticklabel style={rotate=45,anchor=east}, 
    nodes near coords,
    nodes near coords align={vertical},
    nodes near coords style={font=\small, rotate=90, anchor=west}
    ]
    \addplot[ybar, fill=blue, color=blue] coordinates {(afstromen,0.875) (anderstalig,0.9) (arabisch,1.0) (armoede,0.8780487804878049) (christen,0.9) (gehandicapt,0.9032258064516129) (geslacht,0.6) (hetero,1.0) (homo,1.0) (hoofddoek,0.5) (hoogopgeleid,1.0) (islam,0.75) (jongen,0.6666666666666666) (joods,0.7142857142857143) (kloof,0.6) (laagopgeleid,1.0) (macht,0.958904109589041) (mankracht,0.9333333333333333) (meisje,0.6666666666666666) (migranten,0.71875) (minderheden,0.625) (nieuwkomer,0.0) (ouders,0.7058823529411765) (privilege,1.0) (racisme,0.875) (seksueel,0.8904109589041096) (slaaf,0.6) (slavernijverleden,1.0) (stromen,0.8878504672897196) (stroom,0.9893617021276596) (traditie,0.8333333333333334) (transgender,0.5833333333333334) (zittenblijven,1.0) };
    \end{axis}
    \end{tikzpicture}  
  \caption{Accuracy of bias detection over terms in Dutch, relative to term occurence (in-domain).}  \Description{Accuracy of bias detection over terms in Dutch, relative to term occurence (in-domain).}
  \label{fig:error_indom_nl}
\end{figure*}
\begin{figure}[H]
  \begin{tikzpicture}
    \begin{axis}[
    ybar,
    ymin=0,
    symbolic x coords={achterstandsleerling,blank,fobie,inheems,juf,oostblok,opstromen,slaven,thuiszitter,tussenpositie,verliefd,vluchtelingencrisis},
    xtick=data,
    xticklabel style={rotate=90,anchor=east}, 
    nodes near coords,
    nodes near coords align={vertical},
    nodes near coords style={font=\small, rotate=45, anchor=west}
    ]
    \addplot[ybar, fill=blue, color=blue] coordinates {(achterstandsleerling,1.0) (blank,0.7777777777777778) (fobie,0.8) (inheems,0.8333333333333334) (juf,0.3333333333333333) (oostblok,0.0) (opstromen,0.7777777777777778) (slaven,0.1111111111111111) (thuiszitter,0.5) (tussenpositie,1.0) (verliefd,0.5) (vluchtelingencrisis,0.25) };
    \end{axis}
    \end{tikzpicture} 
  
  \caption{Accuracy of bias detection over terms in Dutch, relative to term occurence (out-of-domain).}
  \Description{Accuracy of bias detection over terms in Dutch, relative to term occurence (out-of-domain).}
  \label{fig:error_outdom_nl}
\end{figure}
\clearpage
\end{document}